\documentclass[conference]{IEEEtran}
\IEEEoverridecommandlockouts
\usepackage{cite}
\usepackage{amsmath,amssymb,amsfonts}
\usepackage{algorithmic}
\usepackage{graphicx}
\usepackage{textcomp}

\usepackage{hyperref}
\usepackage{url}

\usepackage{amsmath,graphicx}
\usepackage{multirow}
\usepackage{diagbox}
\usepackage{xcolor}
\usepackage{listings}
\usepackage{url}
\usepackage{colortbl} 
\usepackage{caption}
\usepackage{subfig}

\usepackage{xcolor}
\def\BibTeX{{\rm B\kern-.05em{\sc i\kern-.025em b}\kern-.08em
    T\kern-.1667em\lower.7ex\hbox{E}\kern-.125emX}}

\makeatletter
\newcommand{\newlineauthors}{%
  \end{@IEEEauthorhalign}\hfill\mbox{}\par
  \mbox{}\hfill\begin{@IEEEauthorhalign}
}
\makeatother
\begin{document}

\title{\MakeUppercase{Multimodal Object Detection via Probabilistic a priori Information Integration
}}

\author{\IEEEauthorblockN{1\textsuperscript{st} Hafsa El Hafyani}
\IEEEauthorblockA{\textit{IMT Atlantique} \\
\textit{Lab-STICC, UMR CNRS 6285}\\
Brest F-29238, France \\
hafsa.el-hafyani@imt-atlantique.fr}
\and
\IEEEauthorblockN{2\textsuperscript{nd} Bastien Pasdeloup}
\IEEEauthorblockA{\textit{IMT Atlantique} \\
\textit{Lab-STICC, UMR CNRS 6285}\\
Brest F-29238, France \\
bastien.pasdeloup@imt-atlantique.fr}
%\and
\newlineauthors
\IEEEauthorblockN{3\textsuperscript{rd} Camille Yver}
\IEEEauthorblockA{\textit{InPixal} \\
Rennes, France \\
camille.yver@inpixal.com}
\and
\IEEEauthorblockN{4\textsuperscript{th} Pierre Romenteau}
\IEEEauthorblockA{\textit{InPixal} \\
Rennes, France \\
pierre.romenteau@inpixal.com}
}

\maketitle

\begin{abstract}
Multimodal object detection has shown promise in remote sensing. However, multimodal data frequently encounter the problem of low-quality, wherein the modalities lack strict cell-to-cell alignment, leading to mismatch between different modalities. In this paper, we investigate multimodal object detection where only one modality contains the target object and the others provide crucial contextual information. We propose to resolve the alignment problem by converting the contextual binary information into probability maps. We then propose an early fusion architecture that we validate with extensive experiments on the DOTA dataset.
\end{abstract}

\begin{IEEEkeywords}
Object detection, Multimodal fusion, deep learning, alignment, remote sensing
\end{IEEEkeywords}

\section{Introduction}

Object detection has made remarkable progress thanks to the advances of deep neural detectors. While this thematic is generally developed based on ground level sensors images, applying classical object detection techniques (\cite{liu2016ssd} \cite{redmon2016you}) directly to aerial or space images (e.g. satellite and/or drone images) results in reduced performance. Remote sensing images present more complicated background information, larger scale variance and densely small targets. Therefore, practical applications, %of object detection in remote sensing images, 
such as %urban management and 
military surveillance of areal images leverage multimodal data to enhance object detection capabilities, by incorporating complementary information modelling the structure and context of the objects in the scene. %This type of applications usually aims to use edge device models embedded in sensors or drones to guarantee the device's autonomy.
%images' foreground and background.%, object detection can be improved.

However, when complementary information is collected by multiple sensors or sources (e.g., land cover/use, SAM \cite{kirillov2023segment}), it may present problems of low quality or misalignment %\cite{girard2019noisy} 
with the acquired images, due to, for instance, angular acquisition, leading to potential errors in detection. Figure \ref{fig:roundabout} presents an example of \textcolor{black}{low-quality} %misalignment 
problem of %an RGB image and its 
the contextual information (i.e. masks) acquired by three different sensors. Neither of the contextual information matches perfectly the aerial image. The masks of roundabouts (i.e. left) and vehicles (i.e. right) collected by two different sensors show a shift w.r.t. the aerial image (i.e. center). %collected by a third sensor.
Therefore, it is crucial to present the combination of different modalities in a cohesive manner. Failure to align one modality with the others could lead to increased errors in detection. %(\textcolor{red}{Please refer to Appendix \ref{app:1} for why classical alignment methods fail in this case.})

While showing considerable promise in multimodal tasks (\cite{zhang2020contextual,chen2019object}), the performance of object detectors in mis-aligned multimodal geospatial settings is to be verified. Therefore, this paper first investigates the performance of multimodal data compared to unimodal data for geospatial tasks, since multimodal data do not necessarily improve performance \cite{huang2021makes}. Second, we suggest to resolve the mis-alignment issue by transforming contextual information into probability maps. Third, we examine the proposed alignment approach performance compared to no-alignment settings using the DOTA dataset\cite{xia2018dota}.

The rest of this paper is organized as follows. We introduce the related work in Section \ref{related_work}. Section \ref{contributions} gives a thorough problem description, describes the proposed approach for resolving the alignment issue, and discuss the proposed fusion models. Section \ref{experiments} presents the experimental results and evaluation of the multimodal fusion detectors. Last but not least, in section \ref{conclusion}, we summarize our conclusions and provide directions for future work.

%\textcolor{red}{contrainte du modèle embarqué ==> petit}

%\textcolor{red}{dire que dans le scenario, il y a des metadata qui montrent un effet direct et d'autres qui montrent un effet indirect}
%As a result, we introduce a novel approach for multimodal remote sensing object detection to address the limitations of detecting objects in aerial images by combining a priori context information and object detection. First, to resolve the non-alignment issue, the contextual information is transformed into probability maps, which assigns to each pixel a probability of hosting the object. Then, the object detection model employs a simple one-stream Faster-RCNN as detector with both RGB images and probability maps. Experimental results demonstrate the effectiveness of our proposed approach on the DOTA dataset \cite{xia2018dota}.

%\iffalse
\begin{figure}%[!tb]
    \centering
  \includegraphics[width=0.9\linewidth]{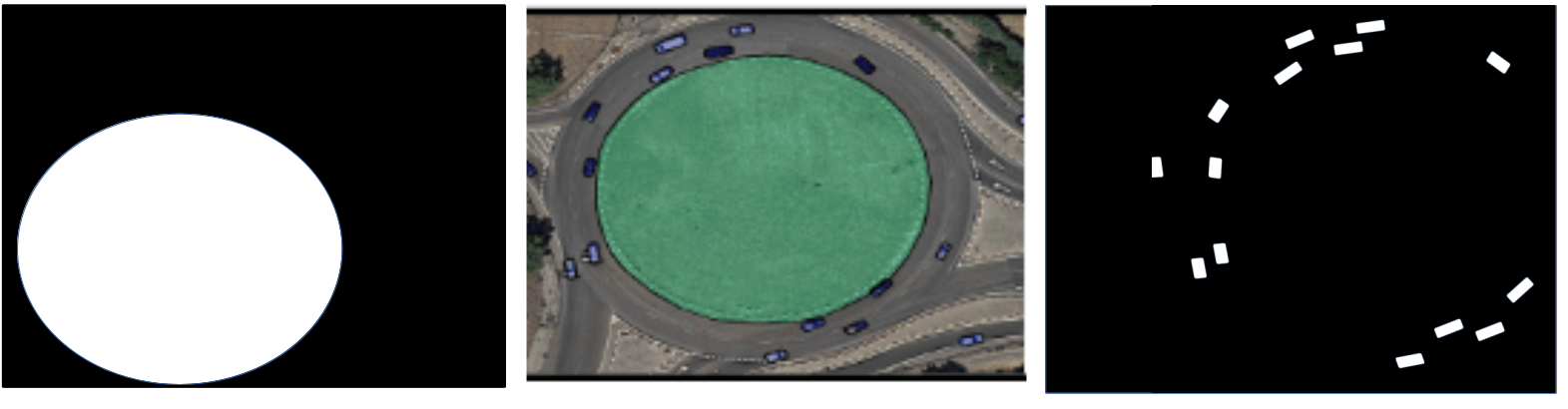}
  % %\vspace{-0.2cm}
  \caption{Example of misalignment of contextual data with an acquired RGB image. On the left, the mask of roundabout does not match the RGB image (center). On the right, the mask of vehicles shows a shift w.r.t. the RGB image.}
  %%\vspace{-0.1cm}
  \label{fig:roundabout}
  % %\vspace{-0.7cm}
\end{figure}
%\fi

\iffalse
\begin{figure*}
\begin{minipage}[c]{0.48\linewidth}
        \centering
        \includegraphics[width=0.6\linewidth]{Figures/roundabout.png}
        \caption{Example of misalignment of contextual data with an acquired RGB image. On the left, the mask of the roundabout does not match the RGB image (center). On the right, the mask of vehicles shows a shift w.r.t. the RGB image.}
        \label{fig:roundabout}
\end{minipage}
\hfill
\hfill
\begin{minipage}[c]{0.48\linewidth}
%\begin{subfigure}[b]{0.45\textwidth}
        \centering
        \begin{subfigure}[b]{0.32\linewidth}
            \centering
            \includegraphics[width=0.6\textwidth]{Figures/test_img.png}
            \caption{Binary mask.}
            \label{fig:example_a}
        \end{subfigure}
        \hfill
        \begin{subfigure}[b]{0.32\linewidth}
            \centering
            \includegraphics[width=0.6\textwidth]{Figures/prob_img.png}
            \caption{Prob map 1.}
            \label{fig:example_b}
        \end{subfigure}
        \hfill
        \begin{subfigure}[b]{0.32\linewidth}
            \centering
            \includegraphics[width=0.6\textwidth]{Figures/new_prob.png}
            \caption{Prob map 2.}
            \label{fig:example_c}
        \end{subfigure}
        \caption{Example of the construction of probability maps from a binary mask (a) based on Eq. \ref{eq:dist1} (b) and Eq. \ref{eq:dist2} (c).}
        \label{fig:example}
   % \end{subfigure}
\end{minipage}%
\end{figure*}
\fi

\section{Related work}\label{related_work}

\subsection{Generic Object Detection Methods}

Deep learning-based algorithms are widely used for object detection with two main categories: two-stage and single-stage approaches. Two-stage methods, such as Faster-RCNN \cite{ren2015faster}, have better detection accuracy as they initially generate candidate boxes, and then identify the object category and refine the object location. One-stage methods, such as SSD \cite{liu2016ssd} and YOLO series \cite{redmon2016you}, offer faster inference speeds since the proposals generation is not needed. Although these methods have shown a good performance on images taken from the ground level, the same performance is not achieved on remote sensing images due to the complex nature of areal images.

On top of that, modern detectors adopt an encoder-decoder architecture based on transformers \cite{dai2022ao2, beal2020toward,maaz2022class}. Specifically, Carion \textit{et al.} \cite{carion2020end} and Zhu \textit{et al.} propose new models called DETR and Deformable DETR, respectively, which have shown competitive results to Faster-RCNN model on large objects. DETR performs less effectively for small objects \cite{carion2020end}. 
Additionaly, Kirillov \textit{et al.} 
unveiled SAM \cite{kirillov2023segment}, a highly accurate pre-trained model for diverse image types, including remote sensing images. Researchers are exploring its potential in handling multimodal data scenarios.

\subsection{Deep Multimodal Fusion}

Researchers have explored various approaches for implementing multimodal fusion to enhance the performance of object detectors in remote sensing. Zhang \textit{et al.} \cite{zhang2020contextual} propose a contextual bidirectional enhancement method to enhance objects' features. Chen \textit{et al.} \cite{chen2019object} propose to combine scene-contextual information in a convolutional neural network to improve the relationship between scene and target. Besides aerial images, several works on the multimodal fusion exist in the literature. Gao \textit{et al.} \cite{gao2020survey} review different approaches to represent multimodal data, and Chu \textit{et al.} \cite{chu2023mt} present MT-DETR, a new fusion model through early and late fusion. Even though these methods have shown good performances when combined with contextual information or other modalities, the alignment problem was not addressed.

\subsection{Image Alignment}

Image alignment is not a new task for the computer vision research community. Brown \cite{brown1992survey} provides a comprehensive review of methods involving geometric transformations to align images based on criteria like similarity. Building upon these approaches, Wang \textit{et al.} \cite{wang2008person} successfully aligned and merged RGB images with thermal images. Girard \textit{et al.} \cite{girard2019noisy} proposed a fully convolutional architecture to correct the alignment of building registries (\emph{e.g.}, from OpenStreetMap) on the image. 
To handle the weakly aligned multispectral data, which makes one object has different positions in different modalities, Zhang \textit{et al.} \cite{zhang2021weakly} proposed a novel Aligned Region CNN (AR-CNN), which includes a region feature alignment module to align the region features. 
following the work of Zhang \textit{et al.} \cite{zhang2021weakly}, Yuan \textit{et al.} \cite{yuan2024improving} propose an end-to-end model called CAGT to solve weak misalignment and fusion imprecision problem in regional level in multispectral data.
Therefore, image alignment has proven successful in these studies, particularly when dealing with two images that depict the same aspects (\emph{e.g.}, RGB images and thermal images) and both exhibit the target objects. However, when attempting to align RGB images with binary masks, the task becomes more and more challenging.

\section{Contributions} \label{contributions}
%\vspace{-0.3cm}
\subsection{Problem Statement}
%\vspace{-0.3cm}

We consider a problem that involves two domains, a \textit{multimodal object detection problem} \cite{gao2020survey} of RGB images and contextual information represented by binary masks retrieved by sensors or segmentation models (e.g. SAM), showing the land cover/use data, 
and \textit{image alignment problem} %(\cite{girard2019noisy,Zhao_2021_CVPR})
(\cite{girard2019noisy, Arar_2020_CVPR, zhao2021deep}) between different modalities. 

Existing work on modality alignment mainly emphasizes datasets where objects are visible across all modalities. However, in our case, objects are exclusively visible within one modality. The alignment of these unique characteristic modalities requires the development of specialized techniques tailored to address this specific circumstance. Traditional approaches may not be directly applicable here, since they are based on the assumption of object presence across multiple modalities. Therefore, innovative solutions are necessitated to  effectively align modalities and fill the gap of modality alignment in the case of the absence of object visibility across all modalities in the dataset. 

Our objective is to train a detector that can combine the multimodal data while ensuring a sound cross-modality representation. Specifically, we desire to obtain an agnostic shift alignment representation that works equally whether the multi modalities are shifted or not. Therefore, we propose a multimodal fusion model for object detection based on probability maps. %The basic idea is to reconstruct probability maps from binary images, which represent the likelihood or confidence the object being present at different locations in the image. 
Indeed, instead of having binary values (\emph{i.e.}, 0 and 1), the probability map assigns a continuous value between 0 and 1 to each pixel, indicating the probability of that pixel hosting the object. \textcolor{black}{Hence, even if the modalities are not perfectly aligned, the probability maps can provide guidance to the model and inform it about the likelihood of the object's presence within the cell.}
%During the training, the model can learn inter- and/or intra-modality interactions, and focus on detecting the target objects while considering the likelihood of actually finding the object.
%\textcolor{red}{Classical alignment methods fail to align the different modalities. Please see Appendix \ref{app:1} for more details.}

%\vspace{-0.4cm}
\subsection{Probability Maps}
%\vspace{-0.2cm}
One straightforward way to generate probabilistic maps %from binary masks 
is the Euclidean distance transformation. Inspired by common sens, it consists of calculating the Euclidean distance of each pixel to the nearest boundary (pixel with a value of 1) in the binary image. As the distance between a pixel and the object's boundary increases, its probability of hosting the object decreases.
Formally, the task involves processing a single channel image represented as a tensor $I_c$ of size $H * W * 1$, where each element $I_c(i,j)$ is either 0 or 1. $H$ and $W$ denote image's height and the width, respectively. We want to compute the probability map $P$ of the same size, where each element $P(i,j)$ represents the probability of the pixel at position $(i, j)$ belonging to the target class:
%First, we compute the normalized distance from each cell to the object:

\vspace{-0.3cm}
\begin{equation}
 \begin{cases} 
    P(i,j) = 1 - d_{norm}(i,j)      \\
    d_{norm}(i,j) = \frac {d(i,j)}{max(d)}    \\
    d(i,j) = min_{i', j' s.t. (i', j') \in I_c} || (i, j) - (i', j') ||_2   \\
  \end{cases} \\
 \label{eq:dist1}
\end{equation}
 %
 %\vspace{0.3cm}

% \vspace{-0.3cm}
% \begin{equation}\label{eq:dist1}
%     P(i,j) = 1 - d_{norm}(i,j)
% \end{equation}

% \vspace{-0.4cm}
% \begin{equation}\label{eq:normalization}
% d_{norm}(i,j) = \frac {d(i,j)}{max(d)}
% % \label{eq:1}
% \end{equation}

where $d_{norm}(i,j)$ is normalized distance from each cell to the object, and $d(i,j)$ is the Euclidean distance of pixel at position $(i,j)$ to the nearest boundary. %defined as :

% \vspace{-0.3cm}
% \begin{equation}\label{eq:euclidean}
%     d(i,j) = min_{i', j' s.t. (i', j') \in I_c} || (i, j) - (i', j') ||_2
% \end{equation}

%Then, we calculate the probability map:

%\vspace{-0.4cm}

% Now the probability map $P$ is generated, and each element $P(i, j)$ represents the probability of the pixel at position $(i, j)$ hosting the target object, with values ranging from 0 to 1. 
%Figure \ref{fig:example} shows an example of the generation of these maps (Figure \ref{fig:example_b}) from a binary mask (Figure \ref{fig:example_a}). The probability is higher for pixels closer to the object boundary and decreases as the distance from the object boundary increases.% as shown in Figure \ref{fig:example_b}. 

%\iffalse
\begin{figure}%[h]
\begin{minipage}[c]{\linewidth}\centering
\subfloat[\label{fig:example_a}]{\includegraphics[width=0.3\linewidth]{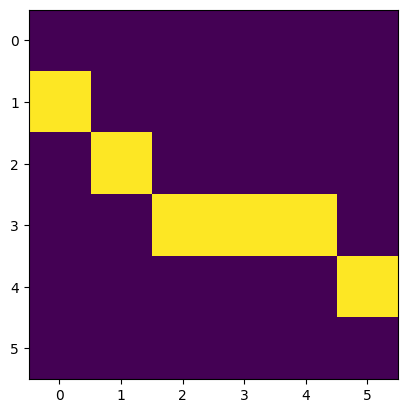}}
\hfil
\subfloat[\label{fig:example_b}]{\includegraphics[width=0.3\linewidth]{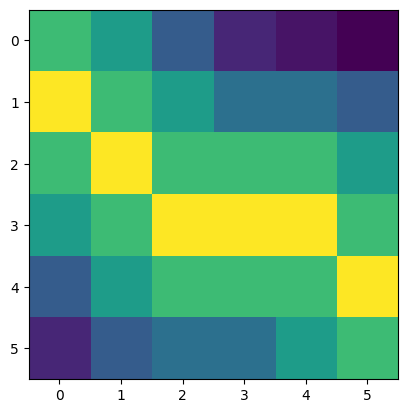}}
\hfil
\subfloat[\label{fig:example_c}]{\includegraphics[width=0.3\linewidth]{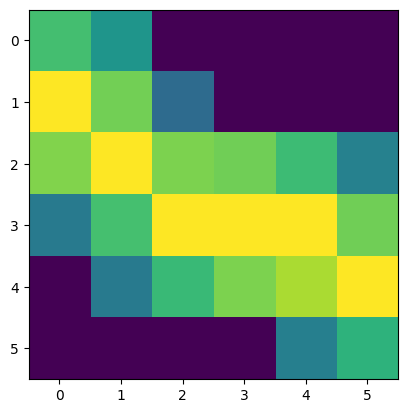}}
    \caption{Example of the construction of probability maps from a binary mask (a) based on Eq. \ref{eq:dist1} (b) and Eq. \ref{eq:dist2} (c).}
    \label{fig:example}
\end{minipage}%
\end{figure}
%\fi

\iffalse
\begin{figure}
     \centering
     \begin{subfigure}[b]{0.32\linewidth}
         \centering
         \includegraphics[width=0.6\textwidth]{Figures/test_img.png}
         \caption{A binary mask.}
         \label{fig:example_a}
     \end{subfigure}
     \hfill
     \begin{subfigure}[b]{0.32\linewidth}
         \centering
         \includegraphics[width=0.6\textwidth]{Figures/prob_img.png}
         \caption{Probability map 1.}
         \label{fig:example_b}
     \end{subfigure}
     \hfill
     \begin{subfigure}[b]{0.32\linewidth}
         \centering
         \includegraphics[width=0.6\textwidth]{Figures/new_prob.png}
         \caption{Probability map 2.}
         \label{fig:example_c}
     \end{subfigure}
     %\vspace{-0.2cm}
        \caption{Example of the construction of probability maps from a binary mask (a) based on Eq. \ref{eq:dist1} (b) and Eq. \ref{eq:dist2} (c).}
        %\vspace{-0.3cm}
        \label{fig:exemple}
        %\vspace{-0.4cm}
\end{figure}
\fi

Figure \ref{fig:example} shows an example of the generation of these maps (Figure \ref{fig:example_b}) from a binary mask (Figure \ref{fig:example_a}). The probability is higher for pixels closer to the object boundary and decreases as the distance from the object boundary increases. As a matter of fact, the above representation of probability maps based on the Euclidean distance shows a smooth propagation of the mask through the whole image. 
%It is primarily optimal when the object is represented by a continuous line in the mask, which is not the case in most real-world applications. 
However, the pixel with the coordinates $(1,1)$ in Figure \ref{fig:example_b} should have a higher probability since it is a neighbour of two cells hosting the object (\emph{i.e.}, cells with coordinates $(0,1)$ and $(1,2)$), as it would be resilient to misalignments along two axes rather than one. Therefore, we propose an alternative distance for constructing the probability maps from the binary masks. It takes into consideration both the distance to the boundaries, and the neighbouring cells as follows: %We formulate the distance as follows:

%\vspace{-0.4cm}
\begin{equation}\label{eq:dist2}
    \forall p \in I_c, \; \; \; d(p) = \frac{
    \sum_{\substack{r \in R(p) \exp(-\alpha d(r,p))
  }} }{\sum_{\substack{m \in M \exp(-\alpha d(m,p))
  }}}
\end{equation}
%\vspace{-0.3cm}

where $R(p)$ is the set of neighbouring cells in a specific radius (to be defined by the user), $\alpha \in \mathbf{R}$ (to be set by the user as well), and $M$ is the set of all the cells containing the mask.
Therefore, with this new representation, the example shown in Figure \ref{fig:example_a} becomes as indicated in Figure \ref{fig:example_c} with $\alpha = 1$ and the radius set to 1 cell in every direction (\emph{i.e.}, left, right, up, down, and diagonal).

%\vspace{-0.5cm}
\subsection{Proposed Method}
%\vspace{-0.2cm}

\begin{figure*}%
    \centering
    %\begin{subfigure}{8cm}
    \includegraphics[width=0.7\linewidth]{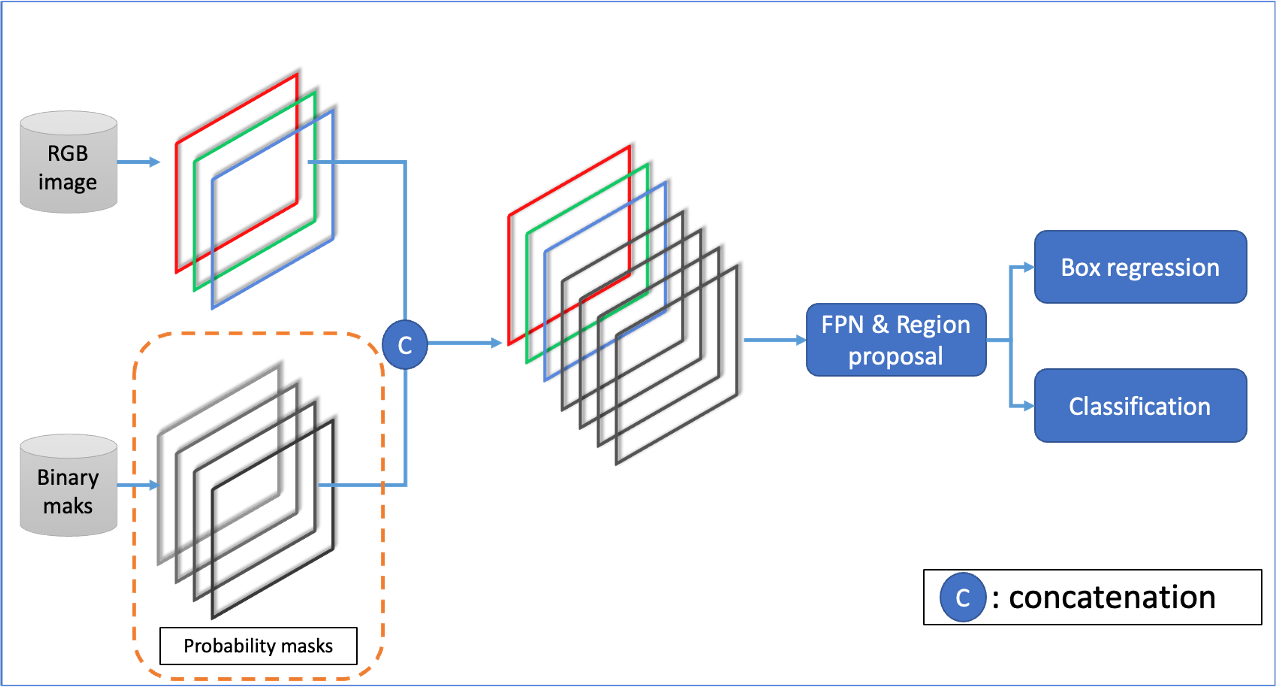}
    %\label{fig:model_direct}
    \caption{The pipeline of our framework.}%
    \label{fig:model_archi}%
    %\vspace{-0.6cm}
\end{figure*}

%Now that we have introduced the concept of probability maps to tackle the misalignment problem, we introduce the flow diagram of our framework. 
The misalignment problem is resolved with the probability maps, we introduce thereafter the flow diagram of our framework. 
As presented in Figure \ref{fig:model_archi}, the framework provides an illustration of the fusion's logic.
%of both direct and indirect fusions. The distinction between both models happens at the level of the dashed orange rectangle.
% consists of two types of fusions: direct and indirect contextual fusion. 
%The direct contextual fusion involves incorporating explicit information about the presence of a specific target object to improve the detection of its bounding boxes. 
%For instance, in the case of detecting buildings in aerial images, including information about the existence of buildings in the images allows the detection model to distinguish the \verb|building| class more effectively than other classes. 
%On the other hand, the indirect fusion model employs fewer contextual data, and entails providing complementary contextual information without directly adding the class itself. This additional context helps the model refine its predictions for the target classes. 
%For example, when detecting vehicles on aerial images, knowing the location of roads and bridges indirectly assists the model in understanding the scene's context and enhances vehicle detection by capturing their spatial relationships with the surrounding structures. 
%The objective of the distinction between direct and indirect fusion is to see to what extend adding direct and indirect contextual information to the detector can improve its performance even if this information is not perfectly aligned.
% Thereafter, the architecture of both models is the same. 
First, we start by generating the probability maps from the binary masks using either equation \ref{eq:dist1} or equation \ref{eq:dist2}. Then, the generated probabilities maps are concatenated to the RGB image, which is then fed to an object detector (e.g. Faster-RCNN). Finally, The detector produces the candidate classes and their bounding boxes with a \textcolor{black}{feature pyramid network (FPN) based network}.

% \textcolor{red}{dire que c'est un dataset artificiel parce qu'on peut pas afficher les données militaires}

%\vspace{-0.3cm}
\section{Experiments} \label{experiments}
%\vspace{-0.3cm}

\subsection{Dataset}
%\vspace{-0.2cm}
\textcolor{black}{Because of the sensibility of military data, we recreate the multi modality miss-alignment problem with a subset of the DOTA dataset \cite{xia2018dota} with its 15 classes as shown in Table \ref{tab:results}.}
% we evaluate our proposed fusion model with probability maps on a subset of the public large-scale dataset for object detection in aerial images (DOTA) \cite{xia2018dota} with its 15 classes as shown in Table \ref{tab:results}.
% DOTA is a geospatial object detection dataset designed for multi-class object detection, comprising 15 different classes. 
%It contains 15 classes shown in Table \ref{tab:results}.
%\verb|plane|, \verb|ship|, \verb|storage tank|, \verb|baseball diamond|, \verb|tennis court|, \verb|basketball court|, \texttt{ground track field}, \verb|harbor|, \verb|bridge|, \verb|large vehicle|, \texttt{small vehicle}, \verb|helicopter|, \verb|roundabout|, \texttt{soccer ball field}, and \verb|swimming pool|. 
%It has 2,806 images and 188,282 instances. The size of the training set, validation set, and testing set are 1/2, 1/6, and 1/3 of the whole dataset, respectively. %A sample of the 15 classes in the DOTA dataset are shown in Figure \ref{fig:dota}. 
The binary masks are extracted from the iSAID dataset \cite{waqas2019isaid}. %We inject a random shift between $5\%$ and $10\%$ of the image width in the binary mask to simulate the misalignment problems. %This dataset has been chosen as the classes allow to test the contextual relationships mentioned in earlier examples.
%\texttt{ground track field}

%\vspace{-0.5cm}
\subsection{Experimental Setup}
%\vspace{-0.2cm}

%In our experimental setup, 
We kept the training set as established by the DOTA dataset. However, we split the validation set between the validation set  and testing set, since the binary masks are only provided for the training and validation sets. Therefore, 2/3 of the DOTA validation set goes to the validation set and 1/3 goes to the test set. %It is worth to note that state-of-the-art object detection \cite{ding2021object} \cite{li2020object} shows enhanced performance capabilities since it uses larger validation and testing sets. 
For the evaluation metrics, we employ the widely used average precision (AP) to evaluate the performance of our model over every class, and the mean average precision (mAP), which is computed over all classes. 
%It is worth to highlight that the assessment of true positives or false positives in a detection task relies on the commonly employed Intersection-Over-Union (IoU) metric. 
Additionally, a candidate bounding box is considered as a true positive if the value of Intersection-Over-Union (IoU) metric is greater than 0.5;
%If the value of IoU is greater than 0.5, the candidate bounding box is considered as a true positive; 
otherwise, it is a false positive. 
All the experiments were made by training a Faster-RCNN \cite{ren2015faster} model, initialized with COCO pre-trained weights. We inject a random shift between $5\%$ and $10\%$ of the image width in the binary mask to simulate the misalignment problems. %We create a random shift between $5\%$ and $10\%$ of the image width in all the binary mask to simulate the misalignment problems. % within the dataset.
%Our implementation is based on the PyTorch framework under Ubuntu 20.04. All the experiments were executed on an NVIDIA GeForce RTX 3090. All the models were trained on 50 epochs, with a batch size of 8, SGD optimizer, weight decay of 5e-4, momentum of 0.9, and the learning rate set to $0.001$. The source code will be publicly available at: \url{https://github.com/helhafyani/Multimodal_fusion}.
Our implementation is based on the PyTorch framework under Ubuntu 20.04. The experiments were executed on NVIDIA GeForce RTX 3090. The models were trained on 50 epochs, with a batch size of 8, SGD optimizer, weight decay of 5e-4, momentum of 0.9, and the learning rate set to $0.001$.
The source code will be publicly available at: \url{https://github.com/helhafyani/Multimodal_fusion}.

%\vspace{-0.4cm}
\subsection{Experimental Results}
%\vspace{-0.2cm}
We perform two types of fusion experiments to evaluate the model's response to the integration of contextual information. The first type involves adding direct contextual information such as the inclusion of shifted, low-quality masks of objects already present in the RGB images (e.g., adding a shifted mask of boats to enhance boat's detection). The second type explores fusion with indirect contextual information, such as shifted masks of elements like harbors to observe their impact on the model's ability to detect specific objects (e.g. boats).

\subsubsection{Does fusion improve detection performance?} %~\\
%\vspace{-0.2cm}

We perform the direct fusion model with 15-class object detection and 15 binary masks \textcolor{black}{(a binary mask for every class)}. %Each mask represents the existence of one class in the image. 
For the indirect fusion model, %we perform a 15-class object detection model with 3 binary masks of \verb|bridges|, \verb|roundabouts|, and \verb|harbors| to introduce a context. %More precisely, 
%In other words, 
we keep the 15 classes in the detection target classes, and use 3 masks: the mask of \verb|harbors| is leveraged for the detection of \verb|ships|, and 
the masks of \verb|bridges| and \verb|roundabouts| are adopted to to enhance the detection of \verb|small| and \verb|large vehicles|. We report the results in Table \ref{tab:results}. \textbf{\textit{Basic}} model refers to the model without fusion. ``\textbf{\textit{Distance in Eq\ref{eq:dist1}}}"  points to the fusion with the probability maps based on  Equation \ref{eq:dist1}, and ``\textbf{\textit{Distance in Eq\ref{eq:dist2}}}" relates to the fusion with probability maps based on Equation \ref{eq:dist2}. %``Direct" and ``Indirect" refer to the direct and indirect fusion models, respectively. 
% We distinguish between the two proposed methods of creating the probability masks. 
\textbf{\textit{Basic fusion}} model is the baseline which refers to the standard fusion model without adopting the probability maps solution. 

\begin{table*}[!htb]
\caption{Performance of the different models compared to the model without fusion (\emph{i.e.}, Basic). Bold entries denote the best detection per target class. Green/orange/red entries show the models that are better/equal/worse compared to the Basic model.}
%\vspace{-0.3cm}
    \label{tab:results}
    \centering
    \resizebox{\linewidth}{!}{%
\begin{tabular}{cc|c|c|c|c|c|c|c|c|c|c|c|c|c|c|c||c|}
% \cline{3-18} 
\hline
\multicolumn{2}{|c|}{\multirow{1}{*}{\textbf{Model}}} &  \textbf{Plane} & \textbf{Ship} & \textbf{ST} & \textbf{BD} & \textbf{TC} & \textbf{BC} & \textbf{GTF} & \textbf{Harbor} & \textbf{Bridge} & \textbf{LV} & \textbf{SV} & \textbf{Heli.} & \textbf{Round.} & \textbf{SBF} & \textbf{SP} & \textbf{mAP}\\
\hline
\multicolumn{2}{|c|}{\multirow{1}{*}{\textbf{Basic}}}  & 0.76& 0.47 & 0.63& 0.62& 0.77& 0.40& 0.39& 0.65& 0.26& 0.59& 0.34& 0.30& 0.46& 0.50& 0.14&0.48 \\ \hline

%\multicolumn{2}{|c|}{\multirow{1}{*}{\textbf{Basic fusion}}}  & 0.78& 0.47 & 0.62& 0.66& 0.77& 0.54& 0.50& 0.62& 0.28& 0.60& 0.33& 0.24& 0.39& 0.53& 0.13& 0.50 \\ \hline

\multicolumn{1}{|c|}{\multirow{2}{*}{\textbf{Basic fusion}}} & 
\textbf{Direct} & \cellcolor{green!25} 0.78& \cellcolor{orange!25} 0.47 & \cellcolor{red!25} 0.62& \cellcolor{green!25} 0.66& \cellcolor{orange!25} 0.77& \cellcolor{green!25} \textbf{0.54}& \cellcolor{green!25} 0.50& \cellcolor{red!25} 0.62& \cellcolor{green!25} 0.28& \cellcolor{green!25} 0.60& \cellcolor{red!25} 0.33& \cellcolor{red!25} 0.24& \cellcolor{red!25} 0.39& \cellcolor{green!25} 0.53& \cellcolor{red!25} 0.13& 0.50 \\ \cline{2-18} 
 
\multicolumn{1}{|c|}{}                       &
\textbf{Indirect} & \cellcolor{orange!25} 0.76& \cellcolor{green!25} 0.48 & \cellcolor{green!25} 0.64& \cellcolor{green!25} 0.66& \cellcolor{green!25} 0.80& \cellcolor{green!25} 0.49& \cellcolor{green!25} 0.50& \cellcolor{red!25} 0.64& \cellcolor{green!25} 0.30& \cellcolor{orange!25} 0.59& \cellcolor{orange!25} 0.34& \cellcolor{orange!25} 0.30& \cellcolor{red!25} 0.42& \cellcolor{orange!25} 0.50& \cellcolor{green!25} 0.15& 0.50 \\  \hline

%%%%%%%%%%%%%

\multicolumn{1}{|c|}{\multirow{2}{*}{\textbf{Distance in Eq.\ref{eq:dist1}}}} & 
\textbf{Direct} & \cellcolor{green!25} \textbf{0.79} & \cellcolor{orange!25} 0.47 & 
 \cellcolor{green!25} \textbf{0.66} & \cellcolor{red!25} 0.60 & \cellcolor{green!25} 0.80 & \cellcolor{green!25} 
 0.44 & \cellcolor{green!25} \textbf{0.62} & \cellcolor{red!25} 0.64 & \cellcolor{green!25} \textbf{0.41} & \cellcolor{red!25} 0.58 & \cellcolor{orange!25} 0.34 & \cellcolor{green!25} \textbf{0.46} & \cellcolor{green!25} \textbf{0.53} & \cellcolor{red!25} 0.48 & \cellcolor{green!25} 0.18 & \textbf{0.53} \\ \cline{2-18} 
 
\multicolumn{1}{|c|}{}                       &
\textbf{Indirect} & \cellcolor{red!25} 0.74 & \cellcolor{green!25} 
0.49 & \cellcolor{orange!25} 0.63 & \cellcolor{green!25} \textbf{0.72} & \cellcolor{green!25} 0.79 & \cellcolor{red!25} 0.37 & \cellcolor{red!25} 0.30 & \cellcolor{red!25} 0.62 & \cellcolor{green!25} 0.38 & \cellcolor{red!25} 0.58 & \cellcolor{red!25} 0.33 & \cellcolor{green!25} 0.38 & \cellcolor{red!25} 0.44 & \cellcolor{green!25} \textbf{0.57} & \cellcolor{green!25} 0.17 & 0.50 \\  \hline

%%%%%

\multicolumn{1}{|c|}{\multirow{2}{*}{\textbf{Distance in Eq.\ref{eq:dist2}}}} & 
\textbf{Direct} & \cellcolor{green!25} 0.78 & \cellcolor{green!25} 0.48 & \cellcolor{red!25} 0.62 & \cellcolor{green!25} 0.67 & \cellcolor{green!25} 0.78 & \cellcolor{green!25} \textbf{0.54} & \cellcolor{green!25} 0.53 & \cellcolor{green!25} \textbf{0.68} & \cellcolor{green!25} 0.35 & \cellcolor{green!25} \textbf{0.61} & \cellcolor{orange!25} 0.34 & \cellcolor{green!25} 0.31 & \cellcolor{green!25} 0.49 & \cellcolor{green!25} 0.55 & \cellcolor{green!25} \textbf{0.21} & \textbf{0.53} \\ \cline{2-18} 

\multicolumn{1}{|c|}{}                       & 
\textbf{Indirect} & \cellcolor{green!25} 0.77 & \cellcolor{green!25} \textbf{0.50} & \cellcolor{red!25} 0.58 & \cellcolor{green!25} 0.67 & \cellcolor{green!25} \textbf{0.81} & \cellcolor{green!25} 0.45 & \cellcolor{green!25} 0.45 & \cellcolor{red!25} 0.62 & \cellcolor{green!25} 0.33 & \cellcolor{red!25} 0.57 & \cellcolor{green!25} \textbf{0.37} & \cellcolor{red!25} 0.22 & \cellcolor{green!25} 0.50 & \cellcolor{green!25} 0.54 & \cellcolor{green!25} 0.16 & 0.50 \\  \hline
%\\ \hline
\end{tabular}}
\end{table*}

%%\vspace{-2cm}
\textcolor{black}{Overall, all of the fusion models perform better than the basic model (\emph{i.e.}, without fusion) when looking at the mAP}, which indicates that adding either direct or indirect contextual information brings along a better distinction between classes. Also, although the direct fusion models show undoubtedly the highest performances, the indirect fusion models show a remarkable performance with the few added contextual information. The indirect model does not only demonstrate a better performance for the related classes, especially \verb|ship| and \verb|large vehicle|, but it facilitates the discrimination between the other classes. %\\
Compared to the direct fusion model (\textit{i.e.} without the probability maps), both direct models perform better, which means adding probability maps improves the detection performance of the models. However, the indirect models have a similar performance in terms of mAP. Still, the models with probability maps perform better in detecting the classes with the contextual information, especially \verb|ship| and \verb|small vehicle|.

\begin{table*}%[!htb]
\caption{Performance of the fusion model with one class at a time. Bold entries denote the best detection per target class. Green/orange/red entries show the classes that are better/equal/worse compared to the Basic model reported in Table \ref{tab:results}.}
%\vspace{-0.3cm}
    \label{tab:fusion_one_class}
    \centering
    \resizebox{\linewidth}{!}{%
\begin{tabular}{|c|c|c|c|c|c|c|c|c|c|c|c|c|c|c|c||c|}
% \cline{3-18} 
\hline
\diagbox[width=10em]{\textbf{Model}}{\textbf{Class}} & \textbf{Plane} & \textbf{Ship} & \textbf{ST} & \textbf{BD} & \textbf{TC} & \textbf{BC} & \textbf{GTF} & \textbf{Harbor} & \textbf{Bridge} & \textbf{LV} & \textbf{SV} & \textbf{Heli.} & \textbf{Round.} & \textbf{SBF} & \textbf{SP} & \textbf{mAP} \\ \hline
% \diag{.05em}{1.2cm}{Model}{Class} & \textbf{Plane} & \textbf{Ship} & \textbf{ST} & \textbf{BD} & \textbf{TC} & \textbf{BC} & \textbf{GTF} & \textbf{Harbor} & \textbf{Bridge} & \textbf{LV} & \textbf{SV} & \textbf{heli} & \textbf{RA} & \textbf{SBF} & \textbf{SP} & \textbf{mAP} \\ \hline
% \textbf{Plane} & 0.78 & 0.48 & 0.59 & 0.62 & 0.78 & 0.44 & 0.50 & 0.63 & 0.31 & 0.58 & 0.33 & 0.23 & 0.40 & 0.45 & 0.11 & 0.48 \\ \hline
\textbf{Plane} & \cellcolor{red!25} 0.75 & \cellcolor{orange!25} 0.47 & \cellcolor{red!25} 0.61 & \cellcolor{green!25} 0.67 & \cellcolor{green!25} 0.78 & \cellcolor{green!25}  \textbf{0.52} & \cellcolor{green!25} 0.51 & \cellcolor{red!25} 0.62 & \cellcolor{green!25} 0.27 & \cellcolor{green!25} 0.61 & \cellcolor{green!25} \textbf{0.35} & \cellcolor{red!25} 0.26 & \cellcolor{red!25} 0.44 & \cellcolor{red!25} 0.47 & \cellcolor{green!25} 0.15 & 0.50 \\ \hline
% \textbf{Ship} & 0.75 & 0.47 & 0.58 & 0.59 & 0.79 & 0.51 & 0.49 & 0.62 & 0.27 & 0.54 & 0.34 & 0.32 & 0.36 & 0.54 &0.15 & 0.49 \\ \hline
\textbf{Ship} & \cellcolor{red!25} 0.75 & \cellcolor{green!25} \textbf{0.49} & \cellcolor{red!25} 0.62 & \cellcolor{orange!25} 0.62 & \cellcolor{green!25} 0.79 & \cellcolor{green!25} 0.43 & \cellcolor{green!25} 0.42 & \cellcolor{red!25} 0.62 & \cellcolor{red!25} 0.24 & \cellcolor{red!25} 0.57 & \cellcolor{red!25} 0.33 & \cellcolor{green!25} 0.33 & \cellcolor{green!25} 0.47 & \cellcolor{green!25} 0.51 & \cellcolor{green!25} 0.19 & 0.49 \\ \hline
\textbf{Storage tank (ST)} & \cellcolor{green!25} 0.78 & \cellcolor{green!25} 0.48 & \cellcolor{red!25} 0.62 & \cellcolor{green!25} 0.64 & \cellcolor{green!25} 0.79 & \cellcolor{green!25} 0.50 & \cellcolor{green!25} 0.47 & \cellcolor{red!25} 0.63 & \cellcolor{green!25} 0.36 & \cellcolor{green!25} 0.60 & \cellcolor{orange!25} 0.34 & \cellcolor{green!25} 0.35 & \cellcolor{green!25} 0.48 & \cellcolor{green!25} 0.53 & \cellcolor{green!25} 0.19 & \textbf{0.52} \\ \hline
\textbf{Baseball diamond (BD)} & \cellcolor{green!25} 0.77 & \cellcolor{green!25} \textbf{0.49} & \cellcolor{red!25} 0.62 & \cellcolor{red!25} 0.60 & \cellcolor{green!25} \textbf{0.81} & \cellcolor{orange!25} 0.40 & \cellcolor{green!25} 0.48 & \cellcolor{orange!25} 0.65 & \cellcolor{green!25} 0.29 & \cellcolor{green!25} 0.60 & \cellcolor{red!25} 0.33 & \cellcolor{green!25} 0.31 & \cellcolor{red!25} 0.44 & \cellcolor{red!25} 0.43 & \cellcolor{green!25} 0.19 & 0.49\\ \hline
\textbf{Tennis court (TC)} & \cellcolor{green!25} 0.78 & \cellcolor{green!25} 0.48 & \cellcolor{red!25} 0.61 & \cellcolor{green!25} 0.64 & \cellcolor{green!25} \textbf{0.81} & \cellcolor{green!25} 0.42 & \cellcolor{green!25} 0.44 & \cellcolor{orange!25} 0.65 & \cellcolor{green!25} 0.29 & \cellcolor{green!25} 0.60 & \cellcolor{orange!25} 0.34 & \cellcolor{green!25} 0.33 & \cellcolor{red!25} 0.39 & \cellcolor{red!25} 0.43 & \cellcolor{green!25} 0.15 & 0.49 \\ \hline
\textbf{Basketball court (BC)} & \cellcolor{green!25} 0.78 & \cellcolor{orange!25} 0.47 & \cellcolor{green!25} 0.64 & \cellcolor{green!25} 0.66 & \cellcolor{orange!25} 0.77 & \cellcolor{green!25} 0.50 & \cellcolor{green!25} 0.48 & \cellcolor{green!25} \textbf{0.66} & \cellcolor{green!25} 0.31 & \cellcolor{orange!25} 0.59 & \cellcolor{orange!25} 0.34 & \cellcolor{green!25} \textbf{0.39} & \cellcolor{red!25} 0.44 & \cellcolor{green!25} \textbf{0.57} & \cellcolor{green!25} 0.18 & \textbf{0.52} \\ \hline
\textbf{Ground track field (GTF)} & \cellcolor{green!25} \textbf{0.79} & \cellcolor{green!25} 0.48 & \cellcolor{orange!25} 0.63 & \cellcolor{green!25} 0.53 & \cellcolor{red!25} 0.75 & \cellcolor{green!25} 0.47 & \cellcolor{green!25} 0.50 & \cellcolor{red!25} 0.60 & \cellcolor{green!25} \textbf{0.35} & \cellcolor{red!25} \cellcolor{green!25} 0.60 & \cellcolor{red!25} 0.33 & \cellcolor{green!25} 0.34 & \cellcolor{red!25} 0.44 & \cellcolor{red!25} 0.45 & \cellcolor{green!25} 0.15 & 0.49\\ \hline
% \textbf{Harbor} & 0.78 & 0.46 & 0.58 & 0.64 & 0.79 & 0.41 & 0.31 & 0.62 & 0.29 & 0.56 & 0.36 & 0.30 & 0.43 & 0.44 & 0.18 & 0.48\\ \hline
\textbf{Harbor} & \cellcolor{green!25} 0.78 & \cellcolor{green!25} 0.48 & \cellcolor{orange!25} 0.63 &  \cellcolor{red!25} 0.57 & \cellcolor{green!25} 0.79 & \cellcolor{green!25} 0.46 & \cellcolor{red!25} 0.32 & \cellcolor{red!25} 0.64 & \cellcolor{green!25} 0.28 & \cellcolor{orange!25} 0.59 & \cellcolor{green!25} \textbf{0.35} & \cellcolor{green!25} 0.35 & \cellcolor{red!25} 0.44 & \cellcolor{red!25} 0.40 & \cellcolor{green!25} 0.19 & 0.49\\ \hline
% \textbf{Bridge} & 0.77 & 0.45 & 0.59 & 0.49 & 0.78 & 0.47 & 0.41 & 0.63 & 0.39 & 0.55 & 0.34 & 0.32 & 0.41 & 0.50 & 0.16 & 0.49\\ \hline
\textbf{Bridge} & \cellcolor{green!25} 0.77 & \cellcolor{green!25} 0.48 & \cellcolor{green!25} \textbf{0.65} &  \cellcolor{red!25} 0.58 & \cellcolor{green!25} 0.80 & \cellcolor{green!25} 0.50 & \cellcolor{green!25} 0.43 & \cellcolor{orange!25} 0.65 & \cellcolor{green!25} 0.34 & \cellcolor{orange!25} 0.59 & \cellcolor{orange!25} 0.34 & \cellcolor{green!25} 0.33 & \cellcolor{red!25} 0.41 & \cellcolor{red!25} 0.47 & \cellcolor{green!25} 0.20 & 0.50\\ \hline
\textbf{Large vehicle (LV)} & \cellcolor{green!25} 0.77 & \cellcolor{orange!25} 0.47 & \cellcolor{green!25} \textbf{0.65} & \cellcolor{green!25} 0.65 & \cellcolor{green!25} 0.79 & \cellcolor{green!25} 0.43 & \cellcolor{red!25} 0.37 & \cellcolor{red!25} 0.63 & \cellcolor{green!25} \textbf{0.35} & \cellcolor{orange!25} 0.59 & \cellcolor{green!25} \textbf{0.35} & \cellcolor{green!25} 0.34 & \cellcolor{red!25} 0.42 & \cellcolor{green!25} 0.53 & \cellcolor{green!25} 0.16 & 0.50\\ \hline
\textbf{Small vehicle (SV)} & \cellcolor{green!25} 0.78 & \cellcolor{green!25} 0.48 & \cellcolor{green!25} 0.64 & \cellcolor{green!25} \textbf{0.72} & \cellcolor{green!25} 0.79 & \cellcolor{green!25} 0.48 & \cellcolor{green!25} 0.40 & \cellcolor{red!25} 0.63 & \cellcolor{green!25} 0.30 & \cellcolor{green!25} \textbf{0.62} & \cellcolor{red!25} 0.32 & \cellcolor{green!25} 0.34 & \cellcolor{green!25} 0.54 & \cellcolor{green!25} 0.51 & \cellcolor{green!25} 0.17 & 0.51 \\ \hline
% \textbf{Helicopter} & 0.79 & 0.46 & 0.55 & 0.67 & 0.78 & 0.48 & 0.34 & 0.64 & 0.31 & 0.56 & 0.32 & 0.20 & 0.32 & 0.43 & 0.12 & 0.46\\ \hline
\textbf{Helicopter} & \cellcolor{green!25} 0.78 & \cellcolor{green!25} \textbf{0.49} & \cellcolor{red!25} 0.60 & \cellcolor{green!25} 0.65 & \cellcolor{green!25} 0.79 & \cellcolor{green!25} 0.46 & \cellcolor{green!25} 0.43 & \cellcolor{orange!25} 0.65 & \cellcolor{green!25} 0.31 & \cellcolor{green!25} 0.60 & \cellcolor{orange!25} 0.34 & \cellcolor{green!25} 0.36 & \cellcolor{red!25} 0.35 & \cellcolor{red!25} 0.48 & \cellcolor{green!25} 0.18 & 0.50\\ \hline
% \textbf{Roundabout} & 0.75 & 0.48 & 0.58 & 0.67 & 0.77 & 0.44 & 0.47 & 0.59 & 0.30 & 0.59 & 0.33 & 0.26 & 0.60 & 0.47 & 0.10 & 0.49\\ \hline
\textbf{Roundabout} & \cellcolor{red!25} 0.72 & \cellcolor{green!25} 0.48 & \cellcolor{red!25} 0.60 & \cellcolor{red!25} 0.58 & \cellcolor{green!25} 0.79 & \cellcolor{green!25} 0.46 & \cellcolor{green!25} 0.47 & \cellcolor{red!25} 0.61 & \cellcolor{green!25} 0.29 & \cellcolor{green!25} 0.61 & \cellcolor{orange!25} 0.34 & \cellcolor{red!25} 0.29 & \cellcolor{green!25} \textbf{0.57} & \cellcolor{red!25} 0.45 & \cellcolor{green!25} 0.15 & 0.50\\ \hline
\textbf{Soccer ball field (SBF)} & \cellcolor{green!25} 0.78 & \cellcolor{green!25} \textbf{0.49} & \cellcolor{orange!25} 0.63 & \cellcolor{orange!25} 0.62 & \cellcolor{green!25} 0.79 & \cellcolor{green!25} 0.42 & \cellcolor{green!25} \textbf{0.52} & \cellcolor{red!25} 0.61 & \cellcolor{green!25} 0.34 & \cellcolor{green!25} 0.61 & \cellcolor{orange!25} 0.34 & \cellcolor{green!25} 0.37 & \cellcolor{red!25} 0.45 & \cellcolor{green!25} 0.52 & \cellcolor{green!25} 0.19 & 0.51\\ \hline
\textbf{Swimming pool (SP)} & \cellcolor{orange!25} 0.76 & \cellcolor{orange!25} 0.47 & \cellcolor{green!25} 0.64 & \cellcolor{red!25} 0.57 & \cellcolor{green!25} \textbf{0.81} & \cellcolor{orange!25} 0.40 & \cellcolor{green!25} 0.47 & \cellcolor{orange!25} 0.65 & \cellcolor{green!25} 0.30 & \cellcolor{green!25} 0.61 & \cellcolor{green!25} \textbf{0.35} & \cellcolor{green!25} 0.38 & \cellcolor{red!25} 0.45 & \cellcolor{orange!25} 0.50 & \cellcolor{green!25} \textbf{0.23} & 0.51 \\ \hline
% \textbf{Plane} & &&&&&&&&&&&&&&& \\
%\hline
\end{tabular}}
%\vspace{-0.5cm}
\end{table*}

%\vspace{-0.3cm}
\subsubsection{Which contextual information improves class discrimination?} %~\\
%\vspace{-0.3cm}

%We investigate the effect of every class on the other classes in Table \ref{tab:fusion_one_class}. 
%In this experiment, 
To answer this question, we only add a single mask as additional input, and report the obtained performance for each class individually in Table \ref{tab:fusion_one_class}. The matrix diagonal can be understood as a direct fusion process with only the target class used as contextual information, while off-diagonal entries show the indirect fusion, \emph{i.e.}, the impact of the contextual class for detecting other classes.

While it appears that adding an input layer for a probability map helps in the detection process in many cases, straightforward conclusions are not so easy. Indeed, the original reasoning behind this work was that presence of a \verb|bridge| or a \verb|roundabout| would probably lead to improved detection performance for \verb|small| and \verb|large vehicles|. However, here, we observe marginal or no improvement at all. The same observation holds for \verb|harbor|, that only brings a marginal gain for detection of \verb|ships|. However, we observe that adding a layer for \verb|planes| brings a 12pt gain for \verb|basketball courts|, or that adding a layer for \verb|small vehicles| improves detection of \texttt{baseball diamonds} by 10pt, which is hard to understand semantically speaking.
Regarding the direct fusion though, we observe that the best performance was achieved for 4 classes out of the 15, and that performance was either improved or stayed the same for 10 of these classes. This suggests again that a priori knowledge of an element in the image improves its detection, even when we have an approximate idea of its location due to misalignments.

%\vspace{-0.5cm}
\section{Conclusion \& perspectives} \label{conclusion}
%\vspace{-0.3cm}

We explore the effectiveness of multimodal data compared to unimodal for geospatial detection. We introduce a solution to mitigate the misalignment issue, involving the transformation of contextual information into probability maps. 
We validate the proposed alignment approach with two fusion type settings using the DOTA dataset. For future work, we intend to explore other fusion methods such as mid-fusion at different layers of the architecture. Additionally, 
we plan to evaluate the performance of the proposed approach on other datasets. 

\section*{Acknowledgment}

This work has been supported by the French National Research Agency (ANR), under the grant agreement ANR-21-ASIA-0004.

\bibliography{IEEEexample}
\bibliographystyle{IEEEtran}

\end{document}